\documentclass[sigconf]{acmart}
\usepackage{booktabs}
\usepackage{multirow}
\usepackage{xcolor}
\usepackage{adjustbox}
\usepackage{threeparttable}
\usepackage{bm}
\usepackage{epsfig}
\usepackage{graphicx}
\usepackage{amsmath}
\usepackage{enumitem} 
\newcommand{\ie}{\textit{i}.\textit{e}.}
\newcommand{\eg}{\textit{e}.\textit{g}.}

\newcommand{\etal}{\textit{et}.\textit{al}.}

\AtBeginDocument{%
  \providecommand\BibTeX{{%
    \normalfont B\kern-0.5em{\scshape i\kern-0.25em b}\kern-0.8em\TeX}}}

\copyrightyear{2021}
\acmYear{2021}
\setcopyright{acmcopyright}\acmConference[HUMA '21]{Proceedings of the 2nd International Workshop on Human-centric Multimedia Analysis}{October 20, 2021}{Virtual Event, China}
\acmBooktitle{Proceedings of the 2nd International Workshop on Human-centric Multimedia Analysis (HUMA '21), October 20, 2021, Virtual Event, China}
\acmPrice{15.00}
\acmDOI{10.1145/3475723.3484247}
\acmISBN{978-1-4503-8671-5/21/10}

\begin{document}
\fancyhead{}

\title{A Closer Look at Temporal Sentence Grounding in Videos: Dataset and Metric}

\author{Yitian Yuan}
\email{yuanyitian@foxmail.com}
\affiliation{
  \institution{Tsinghua University}
  \city{}
  \country{}
}

\author{Xiaohan Lan}
\email{lanxh20@mails.tsinghua.edu.cn}
\affiliation{
  \institution{Tsinghua University}
  \city{}
  \country{}
}

\author{Xin Wang}
\email{xin\_wang@tsinghua.edu.cn}
\affiliation{
  \institution{Tsinghua University \& Pengcheng Laboratory}
  \city{}
  \country{}
}

\author{Long Chen}
\authornote{Corresponding authors. This work started when Long Chen at Tencent.}
\email{zjuchenlong@gmail.com}
\affiliation{
  \institution{Columbia University}
  \city{}
  \country{}
}

\author{Zhi Wang}
\email{wangzhi@sz.tsinghua.edu.cn}
\affiliation{
  \institution{Tsinghua University}
  \city{}
  \country{}
}

\author{Wenwu Zhu$^{*}$}
\email{wwzhu@tsinghua.edu.cn}
\affiliation{
  \institution{Tsinghua University \& Pengcheng Laboratory}
  \city{}
  \country{}
}

\renewcommand{\shortauthors}{Trovato and Tobin, et al.}

\begin{abstract}
Temporal Sentence Grounding in Videos (TSGV), \ie, grounding a natural language sentence which indicates complex human activities in a long and untrimmed video sequence, has received unprecedented attentions over the last few years. Although each newly proposed method plausibly can achieve better performance than previous ones, current TSGV models still tend to capture the moment annotation biases and fail to take full advantage of multi-modal inputs. Even more incredibly, several extremely simple baselines without training can also achieve state-of-the-art performance. In this paper, we take a closer look at the existing evaluation protocols for TSGV, and find that both the prevailing dataset splits and evaluation metrics are the devils to cause unreliable benchmarking. To this end, we propose to re-organize two widely-used TSGV benchmarks (ActivityNet Captions and Charades-STA). Specifically, we deliberately make the ground-truth moment distribution \emph{different} in the training and test splits, \ie, out-of-distribution (OOD) testing. Meanwhile, we introduce a new evaluation metric ``dR@$n$,IoU@$m$'' to calibrate the basic IoU scores by penalizing on the bias-influenced moment predictions and alleviate the inflating evaluations caused by the dataset annotation biases such as overlong ground-truth moments. Under our new evaluation protocol, we conduct extensive experiments and ablation studies on eight state-of-the-art TSGV methods. All the results demonstrate that the re-organized dataset splits and new metric can better monitor the progress in TSGV. Our reorganized datsets are available at \textcolor{blue}{\url{https://github.com/yytzsy/grounding_changing_distribution}}.
\end{abstract}

\ccsdesc[500]{Computing methodologies~Artificial intelligence}
\ccsdesc[300]{Computing methodologies~Natural language processing}
\ccsdesc[300]{Computing methodologies~Computer vision}

\keywords{temporal sentence grounding in videos, dataset bias, evaluation metric, dataset re-splitting, out-of-distribution testing}

\maketitle

\section{Introduction}

\begin{figure}[!t]
	\centering
	\setlength{\abovecaptionskip}{0.0cm}
	\setlength{\belowcaptionskip}{-0.2cm}
	\includegraphics[width=1.0\columnwidth]{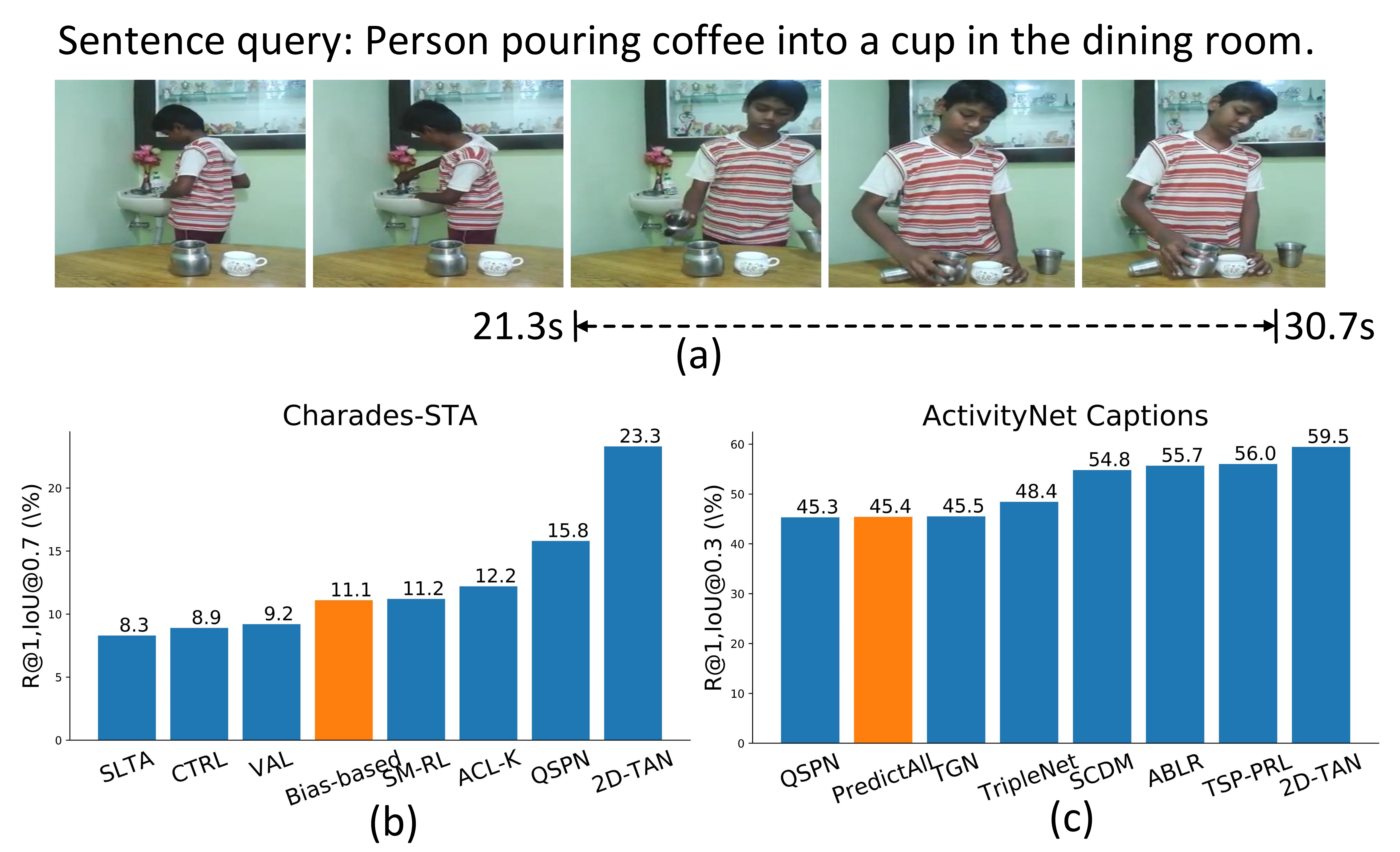}
	\caption{(a): TSGV aims to localize a moment with the start timestamp (21.3s) and end timestamp (30.7s). (b): The performance comparisons of some SOTA TSGV models with Bias-based baseline (orange bar) on Charades-STA with evaluation metric R@$1$,IoU@$0.7$. (c): The performance comparisons of some SOTA TSGV models with PredictAll baseline (orange bar) on ActivityNet Captions with evaluation metric R@$1$,IoU@$0.3$.
	}
	\label{fig:intro}
\end{figure}

Detecting human activities of interest from untrimmed videos is a prominent and fundamental problem in video scene understanding. Early video action localization works~\cite{shou2016temporal,wang2016temporal} mainly focus on detecting activities belonging to some predefined categories~\cite{shou2016temporal,wang2016temporal}, which extremely restrict their flexibility and can hardly cover various human activities in life. For this purpose, a more challenging but meaningful task which extends the limited categories to open natural language descriptions was proposed~\cite{anne2017localizing,gao2017tall,hendricks2018localizing}, dubbed as \textbf{Temporal Sentence Grounding in Videos} (TSGV). As the example shown in Figure~\ref{fig:intro} (a), given a natural language query and an untrimmed video, TSGV needs to identify the start and end timestamps of one segment (\ie, moment) in the video, which semantically corresponds to the language query. Due to its profound significance, the TSGV task has received unprecedented attentions over the last few years --- a surge of datasets~\cite{anne2017localizing,gao2017tall,krishna2017dense,regneri2013grounding} and methods~\cite{chen2018temporally,chen2020rethinking,duan2018weakly,gao2019wslln,ge2019mac,hahn2019tripping,he2019read,jiang2019cross,liu2018attentive,liu2018cross,lu2019debug,wang2019language,wu2020tree,xu2019multilevel,yuan2019semantic,yuan2019find,zeng2020dense,zhang2019man,zhang2020learning,zhang2019cross} have been developed.



Although each newly proposed method can plausibly achieve better performance and make progress over previous ones, a recent study~\cite{otani2020uncovering} shows that even today's state-of-the-art (SOTA) TSGV models still fail to make full use of multi-modal inputs, \ie, they over-rely on the ground-truth moment annotation biases in current benchmarks, and lack of sufficient understanding of the multi-modal inputs. Specifically, taking one prevailing benchmark Charades-STA~\cite{gao2017tall} as an example, suppose that there is a Bias-based baseline model which only makes predictions by sampling a moment from the frequency statistics of the ground-truth moment annotations in the training set. As illustrated in Figure~\ref{fig:intro} (b), this naive Bias-based model unexpected surpasses several SOTA deep models, \ie, Charades-STA has obvious moment location annotation biases. Therefore, \emph{we argue that current TSGV datasets with heavily biased annotations cannot accurately monitor the progress in TSGV research.}


Meanwhile, another characteristic of the ground-truth moment annotations in current TSGV benchmarks is that they usually have relatively long temporal durations. For example, 40\% queries in the ActivityNet Captions dataset refer to a moment occupying over 30\% temporal ranges of the whole input video. These overlong ground-truth moments incidentally make the current evaluation metrics unreliable. Specifically, the most prevalent evaluation metric for today's TSGV is ``R@$n$,IoU@$m$", \ie, the percentage of testing samples which have at least one of the top-$n$ results with IoU larger than $m$. Due to the difficulty of the TSGV task, almost all published TSGV works tend to use a \emph{small} IoU threshold $m$ (\ie, 0.3) for evaluation, especially for challenging datasets (\ie, ActivityNet Captions~\cite{krishna2017dense}). However, \emph{we argue that the metric R@$n$,IoU@$m$ with small $m$ is unreliable for the datasets with overlong ground-truth moment annotations}. For example, a small IoU threshold can be easily achieved by a long duration moment prediction. As an extreme case, a simple baseline which always directly takes the whole input video as the prediction (cf. the PredictAll baseline in Figure~\ref{fig:intro} (c)) can still achieve a SOTA performance under this metric.


In this paper, to help disentangle the effects of ground-truth moment annotation biases, we propose to resplit the widely-used TSGV benchmarks (\ie, Charades-STA and ActivityNet Captions) by changing their ground-truth moment annotation distributions, obtaining two new evaluation benchmarks: \textbf{Charades-CD} (Charades-STA under Changing Distributions) and \textbf{ActivityNet-CD}. These new splits are created by re-organizing all splits (the training, validation and test sets) of original datasets, and the ground-truth moment distributions are deliberately designed \emph{different} in the training and test splits, \ie, out-of-distribution (OOD) testing. To better evaluate models' generalization ability and compare the performance between the OOD samples and the independent and identically distributed (IID) samples, we also maintain a test split with IID samples, denoted as test-iid set (vs. test-ood set). Meanwhile, we propose a more reliable evaluation metric --- dR@$n$, IoU@$m$ --- for small threshold $m$. This metric calibrates the basic IoU scores with the temporal location discrepancy between the predicted and ground-truth moments, which is expected to reduce the influence of moment durations and restraint the inflating evaluations caused by overlong ground-truth moments in the datasets.



To demonstrate the difficulty of our new splits and monitor the progress in TSGV, we evaluate the performance of eight representative SOTA models on our new evaluation protocol. Our key finding is that the performance of most tested models drops significantly when evaluated on the OOD samples (\textit{i.e.}, the test-ood set) compared to the IID samples (\textit{i.e.}, the test-iid set). This finding provides further evidences that existing methods only fit the moment annotation biases, and fail to bridge the semantic gaps between the video contents and natural language queries. Meanwhile, the proposed metric (dR@$n$,Iou@$m$) can effectively reduce the inflating performance caused by the annotation biases when the IoU threshold $m$ is small.


In summary, we make three contributions in this paper:

\begin{itemize}[leftmargin=1em]
\vspace{-0.5em}

	\item We propose new splits of two prevailing TSGV datasets, which are able to disentangle the effects of annotation biases.
	
	\item We propose a new metric: dR@$n$,IoU@$m$, which is more reliable than the existing metrics, especially when IoU threshold is small.

	\item We conduct extensive studies with several SOTA models. Consistent performance gaps between IID and OOD samples have proven that our new evaluation protocol can better monitor the progress in TSGV.
	
\end{itemize}

\section{Related Works}

\subsection{Temporal Sentence Grounding in Videos}

In this section, we coarsely group existing TSGV methods into four categories:

\noindent\textbf{\emph{Two-Stage Methods.}} Early TSGV methods typically solve this problem in a two-stage fashion: They first extract numerous video segment candidates by temporal sliding windows, and then either match the query sentence with these candidates~\cite{anne2017localizing} or fuse query and video segment features to regress the final position, \eg, CTRL~\cite{gao2017tall}, ACL-K~\cite{ge2019mac}, SLTA~\cite{jiang2019cross}, ACRN~\cite{liu2018attentive}, ROLE~\cite{liu2018cross}, VAL~\cite{song2018val} and BPNet~\cite{xiao2021boundary}. To speed up the sliding window processing, Xu~\etal~\cite{xu2019multilevel} proposed QSPN, which injects text features early to generate segment candidates, and helps to eliminate the unlikely segment candidates and increases the grounding accuracy.

\noindent\textbf{\emph{End-to-End Methods.}} Besides the two-stage framework, some other TSGV works seek to solve the grounding problem in an end-to-end manner~\cite{chen2018temporally,yuan2019semantic,yuan2019find,zeng2020dense,zhang2019man,zhang2020learning,zhang2019cross}. Chen~\etal~\cite{chen2018temporally} proposed TGN, which sequentially scores a set of temporal candidates ended at each frame and generates the final grounding result in one single pass. Similarly, ABLR model also processes video sequences via LSTMs~\cite{yuan2019find}, where the start and end timestamps of the predicted segments are regressed from the attention weights yielded by the multi-pass interaction between videos and queries. There are also some works leveraging temporal convolutional networks to solve the TSGV problem. Zhang~\etal~\cite{zhang2019man} presented MAN, which assigns candidate segment representations aligned with language semantics over different temporal locations and scales in hierarchical temporal convolutional feature maps. Yuan~\etal~\cite{yuan2019semantic} introduced the SCDM, where query semantic is used to control the feature normalization between different temporal convolutional layers, making the query-related video activities tightly compose together. Both MAN and SCDM only consider 1D temporal feature maps, while 2D-TAN~\cite{zhang2020learning} models the temporal relations between video segments by a 2D map. In the 2D map, 2D-TAN encodes the adjacent temporal relation, and learns discriminative features for matching video segments with queries.

\noindent\textbf{\emph{RL-based Methods.}} Some recent models also regard the TSGV task as a sequence decision making problem, and resort to Reinforcement Learning (RL) algorithms. Specifically, Wang~\etal~\cite{wang2019language} introduced a semantic matching RL (SM-RL) model by extracting semantic concepts of videos and fusing them with global context features. Then, video contents are selectively observed and associated with the given sentence in a matching-based manner.  Hahn~\etal~\cite{hahn2019tripping} presented TripNet, which uses RL to efficiently localize relevant activity clips in long videos, by learning how to intelligently hop around the video. Wu~\etal~\cite{wu2020tree} formulated a tree-structured policy based progressive RL (TSP-PRL) model to sequentially regulate the predicated temporal boundaries by an iterative refinement process.

\noindent\textbf{\emph{Weakly Supervised Methods.}} Since the ground-truth annotations for the TSGV task are manually consuming, some works start to extend this problem to a weakly supervised scenario where the ground-truth segments are unavailable in the training stage~\cite{duan2018weakly,gao2019wslln,mithun2019weakly,song2020weakly,tan2019wman}. Mithun~\etal~\cite{mithun2019weakly} utilized a latent alignment between video frames and sentence descriptions with Text-Guided Attention (TGA), and TGA was used during the test stage to retrieve relevant moments. Duan~\etal~\cite{duan2018weakly} took the TSGV task as an intermediate step for dense video captioning, and then they established a cycle system and leveraged the captioning loss to train the whole model. Song~\etal~\cite{song2020weakly} presented a multi-level attentional reconstruction network, which leverages both intra- and inter-proposal interactions to learn a language-driven attention map, and can directly rank the candidate proposals at the inference stage.

\subsection{Biases in TSGV Datasets}

A recent work~\cite{otani2020uncovering} also discusses the dataset bias problem in current TSGV benchmarks. The main contribution of~\cite{otani2020uncovering} is to find and analyse the moment annotation biases in previal benchmarks and perform human studies to demonstrate the disagreement among different annotators. In contrast, we propose to re-split these datasets to reduce these ground-truth moment annotation biases, and introduce a new metric to alleviate the inflating performance of SOTA models, \ie, we go one step further to build a more reasonable and reliable evaluation protocol. Meanwhile, we reproduce and analyse eight different SOTA methods from four different categories on both original and new evaluation protocols.



\section{Dataset and Metric Analysis} \label{original_dataset_analysis}

\subsection{Dataset Analysis} \label{dataset_analysis}

So far, there are four available TSGV datasets in our communities: \textbf{DiDeMo}~\cite{anne2017localizing}, \textbf{TACoS}~\cite{regneri2013grounding}, \textbf{Charades-STA}~\cite{gao2017tall} and \textbf{ActivityNet Captions}~\cite{krishna2017dense}. Since both DiDeMo and TACoS have some inherent and obvious disadvantages (\eg, For DiDeMo, the unit interval of annotations is five seconds; for TACoS, the visual scene is restricted in kitchen), dataset Charades-STA and ActivityNet Captions gradually become the mainstream benchmarks for TSGV evaluation~\cite{chen2018temporally,hahn2019tripping,xu2019multilevel,yuan2019semantic,zeng2020dense,zhang2020learning}. The details about these two datasets are as follows:

\noindent\textbf{Charades-STA.} It is built upon the Charades~\cite{sigurdsson2016hollywood} dataset. The average length of videos in Charades is 30 seconds, and each video is annotated with multiple descriptions, action labels, action intervals, and classes of interacted objects. Gao~\etal~\cite{gao2017tall} extended the Charades dataset to the TSGV task by assigning the temporal intervals to text descriptions and matching the common key words in the interval action labels and texts. In the official split~\cite{gao2017tall}, there are 5,338 videos and 12,408 query-moment pairs in the training set, and 1,334 videos and 3,720 query-moment pairs in the test set (cf. Table~\ref{tab:dataset}).

\noindent\textbf{ActivityNet Captions.} It is originally developed for the dense video captioning task~\cite{krishna2017dense}. Since the official test set is withheld, previous TSGV works~\cite{yuan2019semantic,yuan2019find} merge the two available validation subsets ``val1" and ``val2" as the test set. In summary, there are 10,009 videos and 37,421 query-moment pairs in the training set, and 4,917 videos and 34,536 query-moment pairs in the test set. (cf. Table~\ref{tab:dataset}).


\begin{figure}[!t]
	\centering
	\includegraphics[width=1.0\columnwidth]{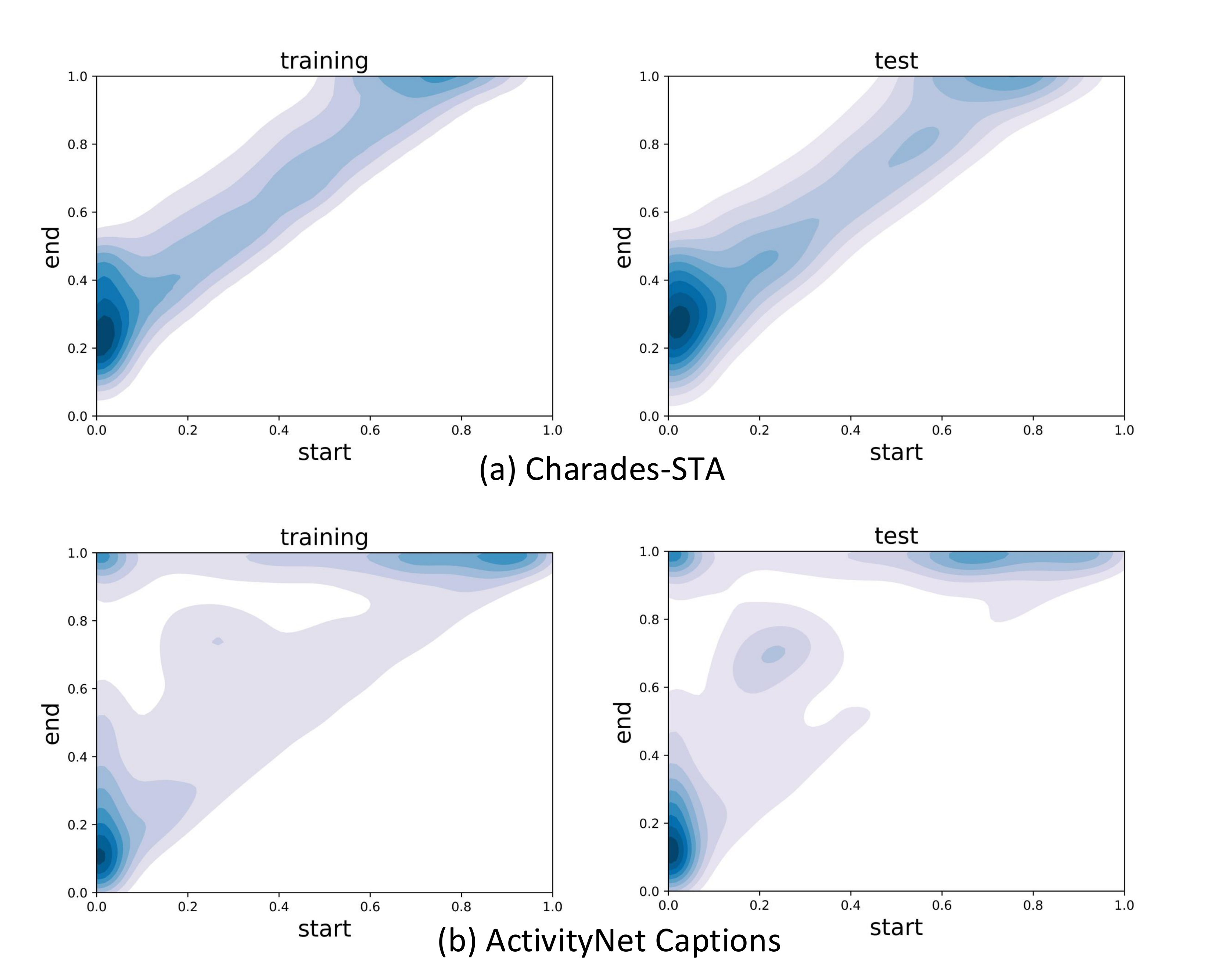}
	\caption{The ground-truth moment annotation distributions of all query-moment pairs in Charades-STA and ActivityNet Captions. The deeper the color, the larger density in distributions.}
	\label{fig:origin_split_dataset}
\end{figure}

To examine the ground-truth moment annotation distributions of these two datasets, we normalize the start and end timestamps of all annotated moments in both the training and test sets, and use Gaussian kernel density estimation to fit the joint distribution of these normalized start and end timestamps. As shown in Figure~\ref{fig:origin_split_dataset}, for both two datasets, the moment annotation distributions are almost identical in the training and test sets. \textbf{For Charades-STA}, most of the moments start at the beginning of the videos and end at around $20\%-40\%$ of the length of the videos. The moment annotation distributions present a strip with relatively uniform width, which indicates that the length of moment in Charades-STA roughly concentrates within a certain range. \textbf{For ActivityNet Captions}, the distributions are significantly different from those of Charades-STA, which concentrates in three local areas, \ie, the three corners. All these areas show that a considerable number of ground-truth moments start at the beginning of the video or end at the ending of the video, even exactly the same as the whole video (the left top area). This may be due to that the dataset ActivityNet Captions is originally annotated for dense video captioning, and the captions (queries) are always annotated based on the whole video. 

Therefore, we can observe that the ground-truth moment annotations in both benchmarks consists of strong biases. In other word, by fitting these moment annotation biases, a simple baseline can also achieve a state-of-the-art performance (cf. Figure~\ref{fig:intro}).

\subsection{Evaluation Metric Analysis} \label{old_metric}

To evaluate the temporal grounding accuracy, almost all existing TSGV works adopt the ``R@$n$,IoU@$m$" as a standard evaluation metric. Specifically, for each query $q_i$, it first calculates the Intersection-over-Union (IoU) between the predicted moment and its ground-truth, and this metric is formally defined as:
\begin{equation}
\text{R@$n$,IoU@$m$} = \frac{1}{N_q} \sum_{i} r(n,m,q_i),
\end{equation}
where $r(n,m,q_i) = 1$ if there is at least one of top-$n$ predicted moments of query $q_i$ having an IoU larger than threshold $m$, otherwise it equals to 0. $N_q$ is the total number of all queries.

\begin{figure}[!t]
	\centering
	\includegraphics[width=1.0\columnwidth]{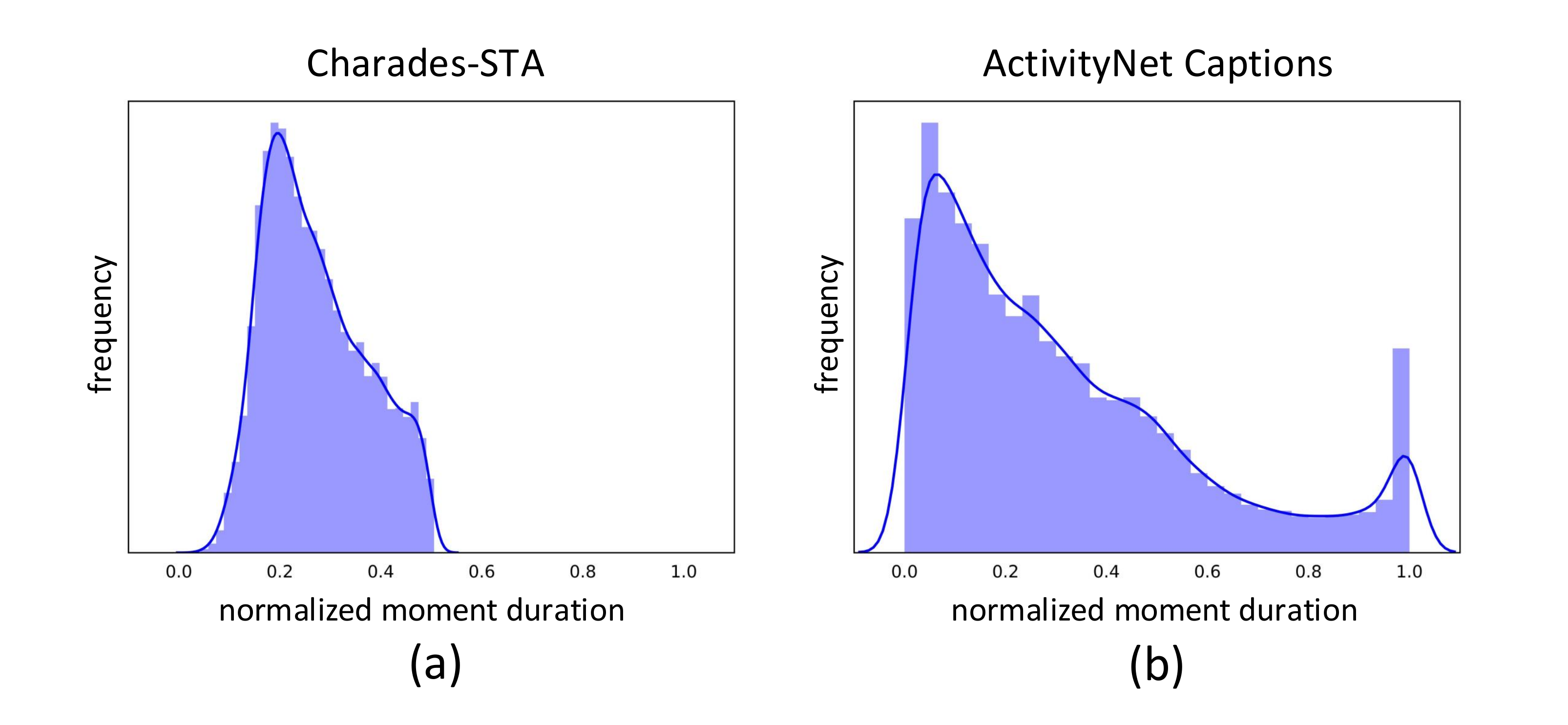}
	\caption{The histogram of the normalized ground-truth moment durations in Charades-STA and ActivityNet Captions.}
	\label{fig:duration}
\end{figure}

Most of previous TSGV methods~\cite{chen2018temporally,liu2018cross,xu2019multilevel,yuan2019find,zhang2020learning} always report their scores on some small IoU thresholds like $m \in \{0.1,0.3,0.5\}$. However, as shown in Figure~\ref{fig:duration} (b), for dataset ActivityNet Captions, a substantial proportion of ground-truth moments have relatively long durations. Statistically, 40\%, 20\%, and 10\% of sentence queries refer to a moment occupying over 30\%, 50\%, and 70\% of the length of the whole video, respectively. Such annotation biases can obviously increase the chance of correct predictions under small IoU thresholds. Taking an extreme case as example, if the ground-truth moment is the whole video, any predictions with duration longer than 0.3 can achieve R@$1$,IoU@$0.3 = 1$. Thus, metric R@$n$,IoU@$m$ with small $m$ is unreliable for current biased annotated datasets.

\begin{figure*}[!t]
	\centering
	\includegraphics[width=1.0\textwidth]{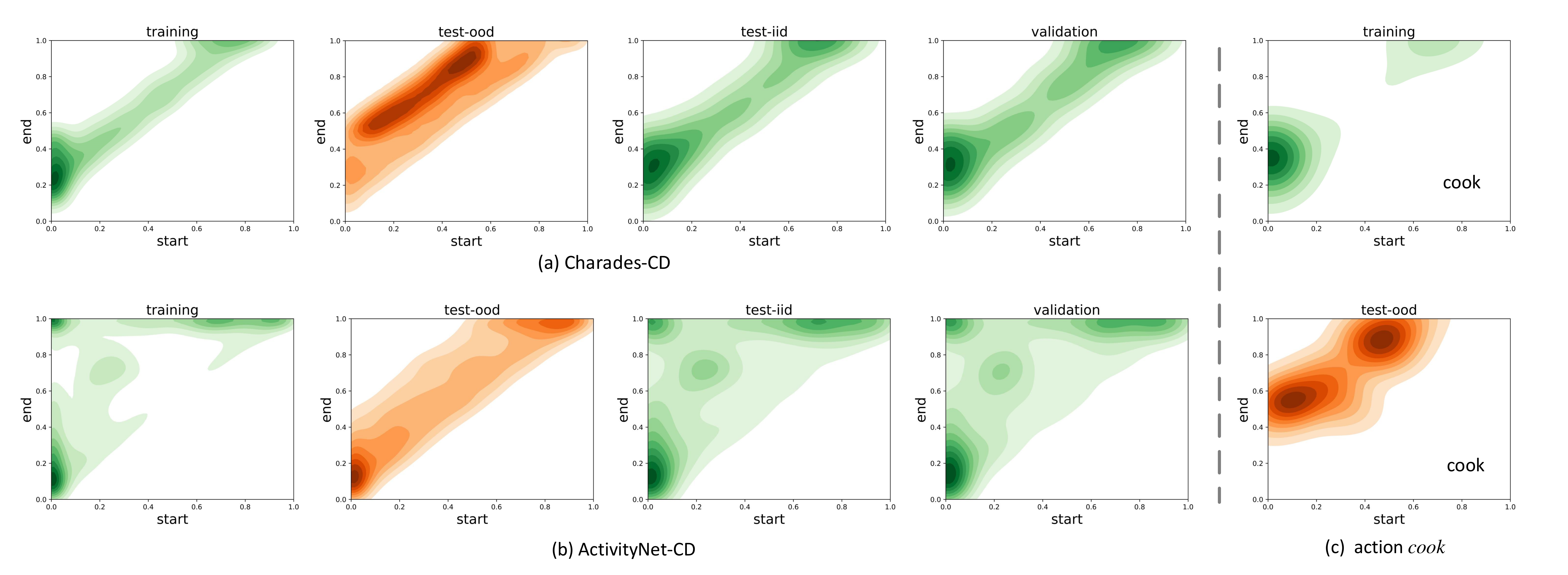}
	\caption{(a) and (b) illustrate the ground-truth moment annotation distributions of each split in Charades-CD and ActivityNet-CD, respectively. (c) presents the moment annotation distributions of the query-indicated moments which contain action \textit{cook} in the training and test-ood sets of Charades-CD. The deeper the color, the larger the density in the distribution.}
	\label{fig:resplit_dataset}
\end{figure*}

\begin{figure}[!t]
	\centering
	\includegraphics[width=1.0\columnwidth]{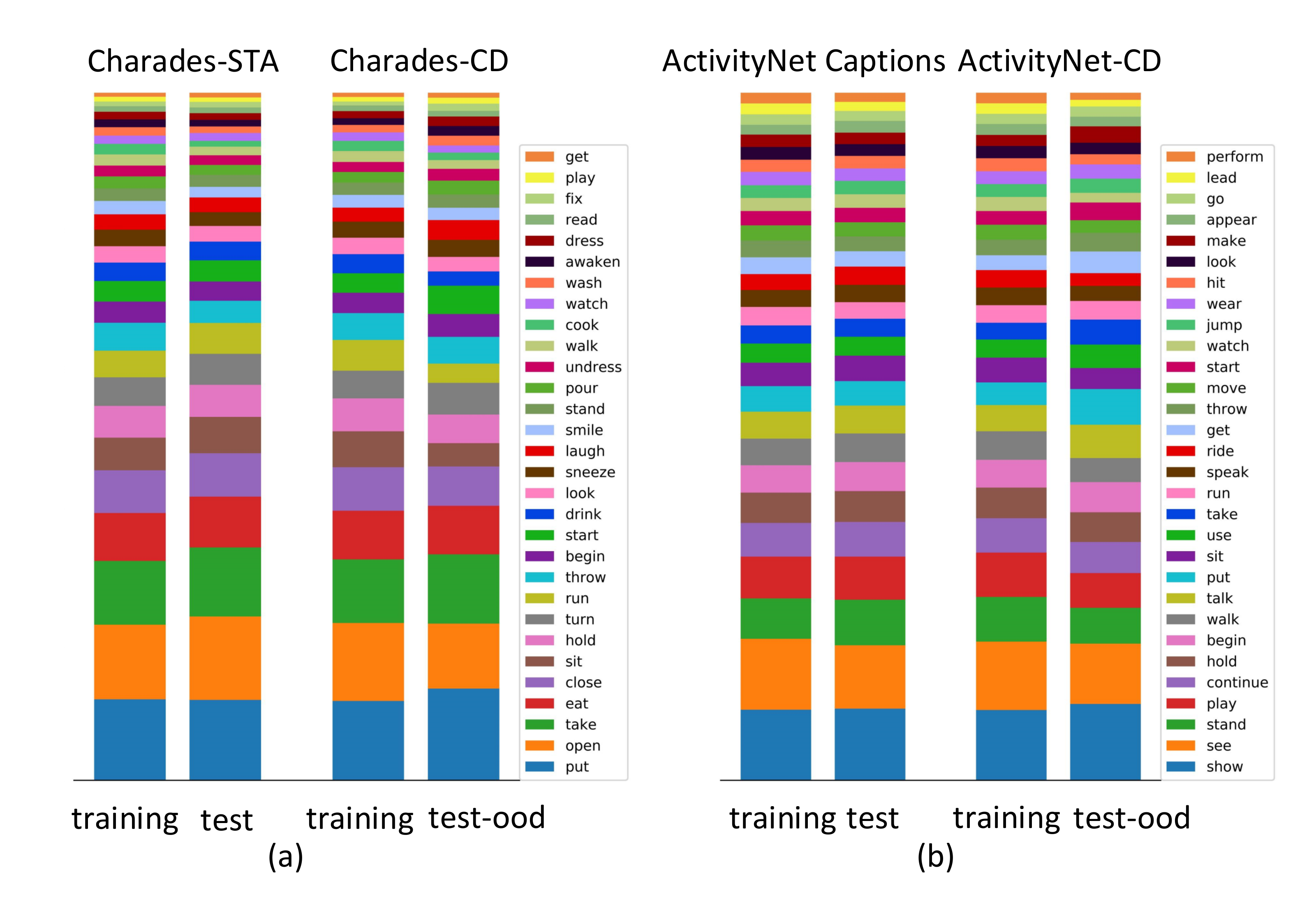}
	\caption{The frequency distributions of the top-30 actions in the query-moment pairs of different splits. The longer the bar, the more frequently the action appears.}
	\label{fig:verb}
\end{figure}

\section{Proposed Evaluation Protocol}

\subsection{Dataset Re-splitting}

To accurately monitor the research progress in TSGV and reduce the influence of moment annotation biases, we propose to re-organize the two datasets (\ie, Charades-STA and ActivityNet Captions) by deliberately assigning different moment annotation distributions in each split. Particularly, each dataset is re-splitted into four sets: \emph{training}, \emph{validation (val)}, \emph{test-iid}, and \emph{test-ood}. All samples in the training, val, and test-iid sets satisfy the independent and identical distribution, and the samples in test-ood set are out-of-distribution. The performance gap between the test-iid set and test-ood set can effectively reflect the generalization ability of the models. We name the two new re-organized datasets as \textbf{Charades-CD} and \textbf{ActivityNet-CD}.

\noindent\textbf{Dataset Aggregation and Splitting.}
For each dataset, we merge the training and test sets by aggregating all the query-moment pairs, \ie, Charades-STA has 12,408 + 3,720 = 16,128 pairs overall and ActivityNet Captions has 37,421 + 34,536 = 71,957 pairs in total (cf. Table~\ref{tab:dataset}). We first regard each query-moment pair as a data sample. Then, we use the Gaussian kernel density estimation as mentioned in Section~\ref{dataset_analysis} (cf. Figure~\ref{fig:origin_split_dataset}) to fit the moment annotation distribution among these data samples. In the fitted distribution, each moment has a density value based on its temporal location in the video. We rank all the moments (as well as their paired queries) based on their density values in a descending order, and take the lower 20\% 
data samples as the preliminary test-ood set, \ie, the temporal locations distribution of the preliminary test-ood set is furthest \emph{different} from the distribution of the whole dataset. The remaining 80\% data sample are divided into the preliminary training set.


\noindent\textbf{Conflicting Video Elimination.} Since each video is associated with multiple sentence queries, another concern is that we need to make sure that there is no video overlap between the training and test sets. Thus, after obtaining the preliminary test-ood set, we check whether the videos of these samples also appear in the preliminary training set. If it is the case, we move all samples (\ie, query-moment pairs) referring to the same video into the split with most of samples. In addition, to avoid the inflating performance of overlong predictions in ActivityNet-CD (cf. the PredictAll baseline in Figure~\ref{fig:intro}), we leave all samples with ground-truth moment occupying over 50\% of the length of the whole video into the training set.

After eliminating all conflicting videos, we obtain the final test-ood set, which consists of around 20\% query-moment pairs of the whole dataset. Then, we randomly divide the remaining samples (based on videos) into three splits: the training, val, and test-iid sets, which consist of around 70\%, 5\%, and 5\% data samples, respectively. The statistics of the new proposed splits are reported in Table~\ref{tab:dataset}.

\begin{table}[t]
    \begin{center}
		\scalebox{1.0}{
			\begin{tabular}{l|lrr}
                \hline
                    Dataset & Split & \# Videos & \# Pairs  \\
                \hline
                    \multirow{2}{*}{Charades-STA} & training    & 5,338     & 12,408  \\
                    & test & 1,334 & 3,720 \\
                \hline
                    \multirow{2}{*}{ActivityNet Captions} & training & 10,009 & 37,421  \\
                    & test & 4,917 & 34,536 \\
                \hline
                    \multirow{4}{*}{Charades-CD} & training & 4,564 & 11,071  \\
                    & val & 333 & 859 \\
                    & test-iid & 333 & 823 \\
                    & test-ood & 1,442 & 3,375 \\
                \hline
                    \multirow{4}{*}{ActivityNet-CD} & training  & 10,984    & 51,415 \\
                    & val & 746 & 3,521 \\
                    & test-iid & 746 & 3,443 \\
                    & test-ood & 2,450 & 13,578 \\
                \hline
			\end{tabular}
		} 
	\end{center}
	\caption[]{The detailed statistics of the number of videos and query-moment pairs in different datasets and splits.}
	\label{tab:dataset}
\end{table}

\subsection{Charades-CD and ActivityNet-CD}

\noindent\textbf{Moment Annotation Distributions.} The ground-truth moment annotation distributions of Charades-CD and ActivityNet-CD are illustrated in Figure~\ref{fig:resplit_dataset}. From the Figure~\ref{fig:resplit_dataset}, we can observe that the moment annotation distributions of the test-ood set are significantly different from those of the other three sets (\ie, training, val, and test-iid sets). Compared with the moment annotation distributions of original test split (cf. Figure~\ref{fig:origin_split_dataset}), the proposed test-ood split has several improvements: 1) For Charades-CD, the distributions of the start timestamps of the moments are more diverse (vs. concentrating on the beginning of the videos). 2) For ActivityNet-CD, more moments locate in relatively central areas of the videos, \ie, models will not perform well by over relying on the annotation biases.


\noindent\textbf{Action Distributions.} We also investigate the action distributions of the original and re-organized datasets. Specifically, for each dataset, we extract the verbs from all sentence queries and count the frequency of each verb. Since the verb frequencies satisfy a long-tail distribution, we select the top-30 frequent verbs, which cover 92.7\% of all action types in Charades-CD and 52.9\% for ActivityNet-CD, respectively. The statistical results are illustrated in Figure~\ref{fig:verb}. From this figure, we can observe that the new test-ood sets on both two datasets still have similar action distributions with the training set and original splits, which shows the OOD of moment annotations comes from each verb type. As shown in Figure~\ref{fig:resplit_dataset} (c), the moment annotation distribution of the new training and test-ood set are totally different for the verb \emph{cook}. 

\subsection{Proposed Evaluation Metric}

\begin{figure}[!t]
	\centering
	\includegraphics[width=1.0\columnwidth]{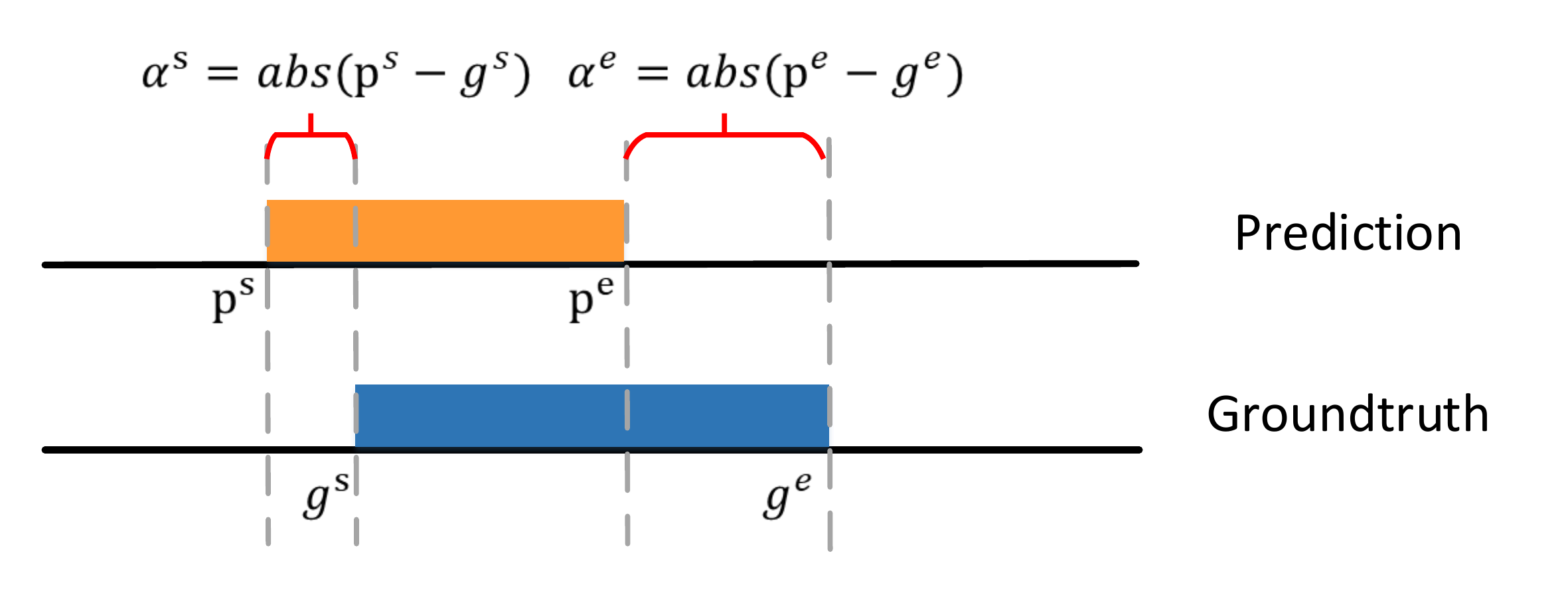}
	\caption{An illustration of the proposed dR@$n$,IoU@$m$ metric.}
	\label{fig:newmetric}
\end{figure}

As discussed in Section~\ref{old_metric}, the most prevailing evaluation metric --- R@$n$,IoU@$m$ --- is unreliable under small threshold $m$. To alleviate this issue, as shown in Figure~\ref{fig:newmetric}, we propose to calibrate the $r(n,m,q_i)$ value by considering the ``temporal distance'' between the predicted and ground-truth moments. Specifically, we propose a new metric discounted-R@$n$,IoU@$m$ (dR@$n$,IoU@$m$):
\begin{equation}
\label{eq:new_metric}
\text{dR@$n$,IoU@$m$} = \frac{1}{N_q}  \sum_{i}r(n,m,q_i) \cdot \alpha^s_i \cdot \alpha^e_i, 
\end{equation}
where $\alpha^*_i =  1-\text{abs}(p_i^*-g_i^*)$, and $\text{abs}(p_i^*-g_i^*)$ is the absolute distance between the boundaries of predicted and ground-truth moments. Both $p_i^*$ and $g_i^*$ are normalized to the range (0, 1) by dividing the whole video length. When the predicted and ground-truth moments are very close to each other, the discount ratio $\alpha^*_i$ will be close to 1, \ie, the new metric can degrade to R@$n$,IoU@$m$ with exactly accurate predictions. Otherwise, even the IoU threshold condition is met, the score $r(n,m,q_i)$ will still be discounted by $\alpha^*_i$, which helps to alleviate the inflating recall scores under small IoU thresholds. With the proposed dR@$n$,IoU@$m$ metric, those speculation methods which over-rely on moments annotation biases (\eg, long moments annotations in ActivityNet Captions) will not perform well.




\begin{figure*}[!t]
	\centering
	\includegraphics[width=1.0\textwidth]{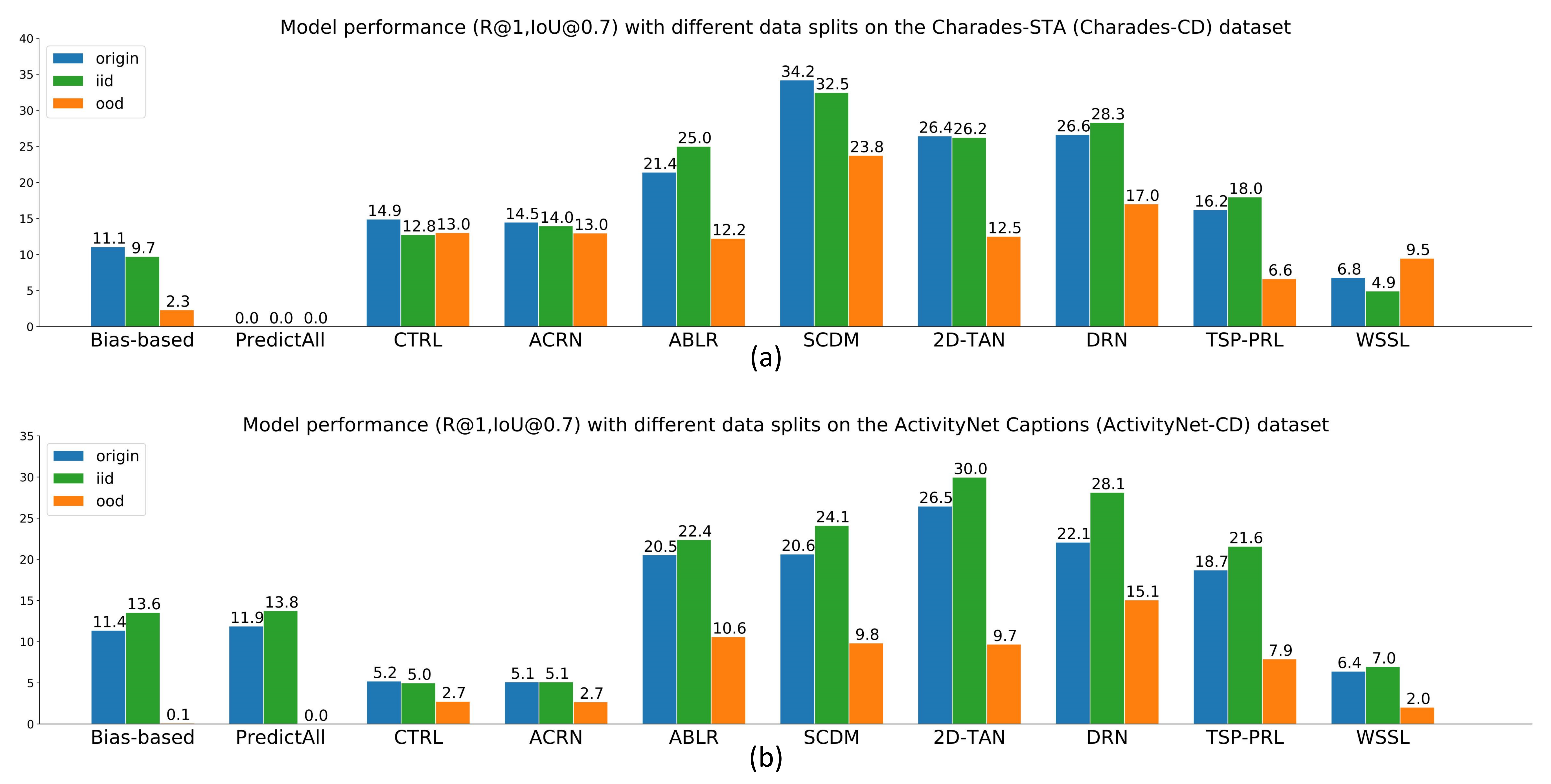}
	\caption{Performances (\%) of SOTA TSGV methods on the test set of original splits (Charades-STA and ActivityNet Captions) and test sets (test-iid and test-ood) of proposed splits (Charades-CD and ActivityNet-CD). We use metric R@$1$,IoU@$0.7$ in all cases.}
	\label{fig:ood_origin_compare}
\end{figure*}

\section{Experiments}

\subsection{Benchmarking the SOTA TSGV Methods}

To demonstrate the difficulty of the new proposed splits (\ie, Charades-CD and ActivityNet-CD), we compare the performance of two simple baselines and eight representative state-of-the-art methods on both the original and proposed splits. Specifically, we can categorize these methods into the following groups: 
\begin{itemize}
\vspace{-0.1em}
\item \emph{Bias-based Method}: It uses the Gaussian kernel density estimation to fit the moment annotation distribution, and randomly samples several locations based on the fitted distribution as the final moment predictions.

\item \emph{PredictAll Method}: It directly predicts the whole video as the final moment predictions.

\item \emph{Two-Stage Methods}: Cross-modal Temporal Regression Localizer (\textbf{CTRL})~\cite{gao2017tall}, and Attentive Cross-modal Retrieval Network (\textbf{ACRN})~\cite{liu2018attentive}.

\item \emph{End-to-End Methods}: Attention-Based Location Regression \\ (\textbf{ABLR})~\cite{yuan2019find}, Semantic Conditioned Dynamic Modulation (\textbf{SCDM})~\cite{yuan2019semantic}, 2D Temporal Adjacent Network (\textbf{2D-TAN})~\cite{zhang2020learning}, and Dense Regression Network (\textbf{DRN})~\cite{zeng2020dense}.

\item \emph{RL-based Method}: Tree-Structured Policy based Progressive Reinforcement Learning (\textbf{TSP-PRL})~\cite{wu2020tree}.

\item \emph{Weakly-supervised Method}: Weakly-Supervised Sentence Localizer (\textbf{WSSL})~\cite{duan2018weakly}.

\end{itemize}


For all these SOTA methods, we use the public available official implementations to get the temporal grounding results. The results of the proposed test-iid and test-ood sets on two datasets come from the same model finetuned on the val set. For more fair comparisons, we have unified the feature representations of the videos and sentence queries. To cater for most of TSGV methods, we use I3D feature~\cite{carreira2017quo} for the videos in dataset Charades-STA (Charades-CD), and C3D feature~\cite{tran2015learning} for the videos in dataset ActivityNet Captions (Activity-CD). Each word in the query sentences is encoded by a pretrained GloVe~\cite{pennington2014glove} word embedding.

\begin{table*}[t]
    \begin{center}
		\scalebox{1.0}{
			\begin{tabular}{ll|ccccc|ccccc}
                \hline
                    \multirow{2}{*}{Method} &\multirow{2}{*}{Split} &\multicolumn{5}{c|}{Charades-CD} &\multicolumn{5}{c}{ActivityNet-CD} \\
                    & & m=0.1 & m=0.3 & m=0.5 & m=0.7 & m=0.9 & m=0.1 & m=0.3 & m=0.5 & m=0.7 & m=0.9 \\
                    \hline
                    
                    \multirow{2}{*}{Bias-based} 
                    &test-iid 
                    &31.42 &26.25 &16.87 &9.34 &2.70 
                    &36.15 &29.31 &19.81 &12.27 &7.68 \\
    			    &test-ood 
    			    &14.75 &9.30 &5.04 &2.21 &0.55 
    			    &21.89 &9.21 &0.26 &0.11 &0.03 \\
    			    \hline
        			
        			\multirow{2}{*}{PredictAll} 
        			&test-iid 
        			&31.04 &10.93 &0.00 &0.00 &0.00 
        			&36.43 &29.62 &20.05 &12.45 &7.83 \\
        			&test-ood 
        			&37.43 &27.13 &0.06 &0.00 &0.00 
        			&21.87 &9.01 &0.00 &0.00 &0.00 \\
    			    \hline
    			    
        			
        			\multirow{2}{*}{CTRL \cite{gao2017tall}} 
        			&test-iid 
        			&50.61 &42.65 &29.80 &11.86 &1.41 
        			&27.34 &19.42 &11.27 &4.29 &0.25 \\
        			&test-ood 
        			&52.80 &44.97 &30.73 &11.97 &1.12 
        			&26.23 &15.68 &7.89 &2.53 &0.20 \\
    			    \hline
    			    
        			
        			\multirow{2}{*}{ACRN \cite{liu2018attentive}} 
        			&test-iid 
        			&53.22 &47.50 &31.77 &12.93 &0.71 
        			&27.69 &20.06 &11.57 &4.41 &0.75 \\
            		&test-ood 
            		&53.36 &44.69 &30.03 &11.89 &1.38 
            		&27.03 &16.06 &7.58 &2.48 &0.17 \\
    			    \hline
    			    
        			
        			\multirow{2}{*}{ABLR \cite{yuan2019find}} 
        			&test-iid 
        			&59.26 &52.26 &41.13 &23.50 &3.66 
        			&55.62 &46.86 &35.45 &20.57 &6.32 \\
        			&test-ood 
        			&54.09 &44.62 &31.57 &11.38 &1.39 
        			&46.88 &33.45 &20.88 &10.03 &2.31 \\
    			    \hline
    			    
        			\multirow{2}{*}{SCDM~\cite{yuan2019semantic}} 
        			&test-iid 
        			&62.47 &58.14 &47.36 &30.79 &6.62 
        			&55.15 &46.44 &35.15 &22.04 &6.07 \\
        			&test-ood 
        			&59.08 &52.38 &41.60 &22.22 &3.81 
        			&45.08 &31.56 &19.14 &9.31 &1.94 \\
    			    \hline
    			    
        			\multirow{2}{*}{2D-TAN~\cite{zhang2020learning}} & test-iid 
        			&59.80 &53.71 &43.46 &24.99 &6.95 
        			&57.11 &49.18 &39.63 &27.36 &9.00 \\
        			&test-ood 
        			&50.87 &43.45 &30.77 &11.75 &1.92 
        			&44.37 &30.86 &18.38 &9.11 &2.05 \\
    			    \hline
    			    
        			
        			\multirow{2}{*}{DRN~\cite{zeng2020dense}} 
        			
        			&test-iid 
        			&57.03 &51.35 &41.91 &26.74 &6.46 
        			&56.96 &48.92 &39.27 &25.71 &6.81 \\
        			&test-ood 
        			&49.17 &40.45 &30.43 &15.91 &3.13 
        			&47.50 &36.86 &25.15 &14.33 &3.76 \\
    			    \hline
    			    
        			
        			\multirow{2}{*}{TSP-PRL~\cite{wu2020tree}}
        			
        			&test-iid 
        			&54.60 &46.44 &35.43 &17.01 &3.57 
        			&53.98 &44.93 &33.93 &19.50 &4.79 \\
        			&test-ood 
        			&42.21 &31.93 &19.37 &6.20 &1.16 
        			&44.23 &29.61 &16.63 &7.43 &1.46 \\
    			    \hline
    			    
        			
        			\multirow{2}{*}{WSSL~\cite{duan2018weakly}}
        			&test-iid 
        			&45.90 &34.99 &14.06 &4.27 &0.00 
        			&36.67 &26.06 &17.20 &6.16 &1.24 \\
        			& test-ood 
        			&49.92 &35.86 &23.67 &8.27 &0.06 
        			&30.71 &17.00 &7.17 &1.82 &0.17 \\
                    \hline
                    
			\end{tabular}
		} 
	\end{center}
	\caption[]{Performances (\%) of SOTA TSGV methods on the Charades-CD and ActivityNet-CD datasets with metric dR@$1$,IoU@$m$.}
	\label{tab:performance_drecall}
\end{table*}

\subsection{Performance Comparisons on the Original and Proposed Data Splits}

We report the performance of all mentioned TSGV methods with metric R@$1$,IoU@$0.7$ in Figure~\ref{fig:ood_origin_compare}. From Figure~\ref{fig:ood_origin_compare}, we can observe that almost all methods have a significant performance gap between the test-iid and test-ood sets, \textit{i.e.}, these methods always over-rely on the moment annotation biases, and fail to generalize to the OOD testing. Meanwhile, the performance results on the original test set and the proposed test-iid set are relatively close, which shows that the moment distribution of the test-iid set is similar to the majority of the whole dataset. We provide more detailed experimental result analyses in the following:

\noindent\textbf{Baseline Methods.} After changing the moment annotation distributions in different splits, the Bias-based method cannot take advantage of the annotation biases and its performance degrades from 13.6\% on the test-iid set to 0.1\% on the test-ood set of ActivityNet-CD. For the PredictAll method, since all the ground-truth moments in Charades-CD are less than 50\% range of the whole videos, naively predicting the whole video as the grounding results will inevitably cause the R@$1$,IoU@$0.7$ scores to 0.0 on this dataset. Since the ground-truth moments in ActivityNet-CD are much longer, the PredictAll method achieves high results at 11.9\% and 13.8\% on the original test set and new test-iid set, respectively. However, in the test-ood set where the longer segments are excluded, the PredictAll method also degrades its performance to 0.0.

\noindent\textbf{Two-Stage Methods.} We find that the two-stage methods (\ie, CTRL and ACRN) are less sensitive to the domain gaps between the test-iid and test-ood sets. This is due to that they use a sliding-window strategy to retrieve video moment candidates, and compare these moment candidates with each query sentence individually. In this manner, all moment candidates are independent to the overall video contents, and the moment annotation distributions have less influence on the model performance. We can also observe that the performance of these two methods on the test-ood set of ActivityNet-CD presents a more obvious drop compared to the performance on test-iid set. In contrast, the performance on the test-iid and test-ood sets of Charades-CD are competing. The main reason is that the ground-truth moments in the test-ood set of Charades-CD always occupy a longer range over the whole videos (cf. Figure~\ref{fig:resplit_dataset} (a), which makes the sliding windows have more chance to hit the ground-truth moments. In summary, although CTRL and ACRN are less sensitive to the moment annotation biases, their grounding performances are still far behind other types of SOTA methods, \eg, SCDM and DRN.

\noindent\textbf{End-to-End Methods.} As for the end-to-end methods (\ie, ABLR, SCDM, 2D-TAN and DRN), we can observe that their performances all drop significantly on the test-ood set compared to the test-iid set on both two datasets. These methods all have considered the whole video contexts and temporal information. 
The initial intention for this design is that numerous queries often contain some words referring to temporal orders and locations such as ``before'', ``after'', ``begin'' and ``end'', or they want to encode the important temporal relations between video moments. Unfortunately, although our test-ood split does not break any video temporal relations, their performance on the OOD testing still drop significantly. This demonstrates that current methods do not play their advantages and fail to utilize the video temporal relation or vision-language interaction for TSGV.


\noindent\textbf{RL-based Method.} The RL-based method TSP-PRL also suffers from obvious performance drops on the test-ood set compared to the test-iid set. Actually, TSP-PRL adopts IoU between the predicted and ground-truth moment as the training reward in the RL framework. In this case, the temporal annotations directly affect the model learning, and the changes of moment annotation distributions will inevitably influence the model performance.

\noindent\textbf{Weakly-supervised Method.} The results of the weakly-supervised method WSSL is \emph{thought-provoking}: it achieves better performance on test-ood set compared to test-iid set in Charades-CD, but results of these splits in ActivityNet-CD are exactly the reverse. After carefully checking the predicted moment results, we find that the normalized (start, end) moment predictions on both two datasets converge on several certain predictions (\ie, (0, 1), (0, 0.5), (0.5, 1)). These results indicate that the WSSL method does not learn to align the video and sentence semantics at all. Instead, it only speculatively guesses several possible locations.



\subsection{Performance Evaluation with dR@$n$,IoU@$m$ }

\begin{figure}[!t]
	\centering
	\includegraphics[width=1.0\columnwidth]{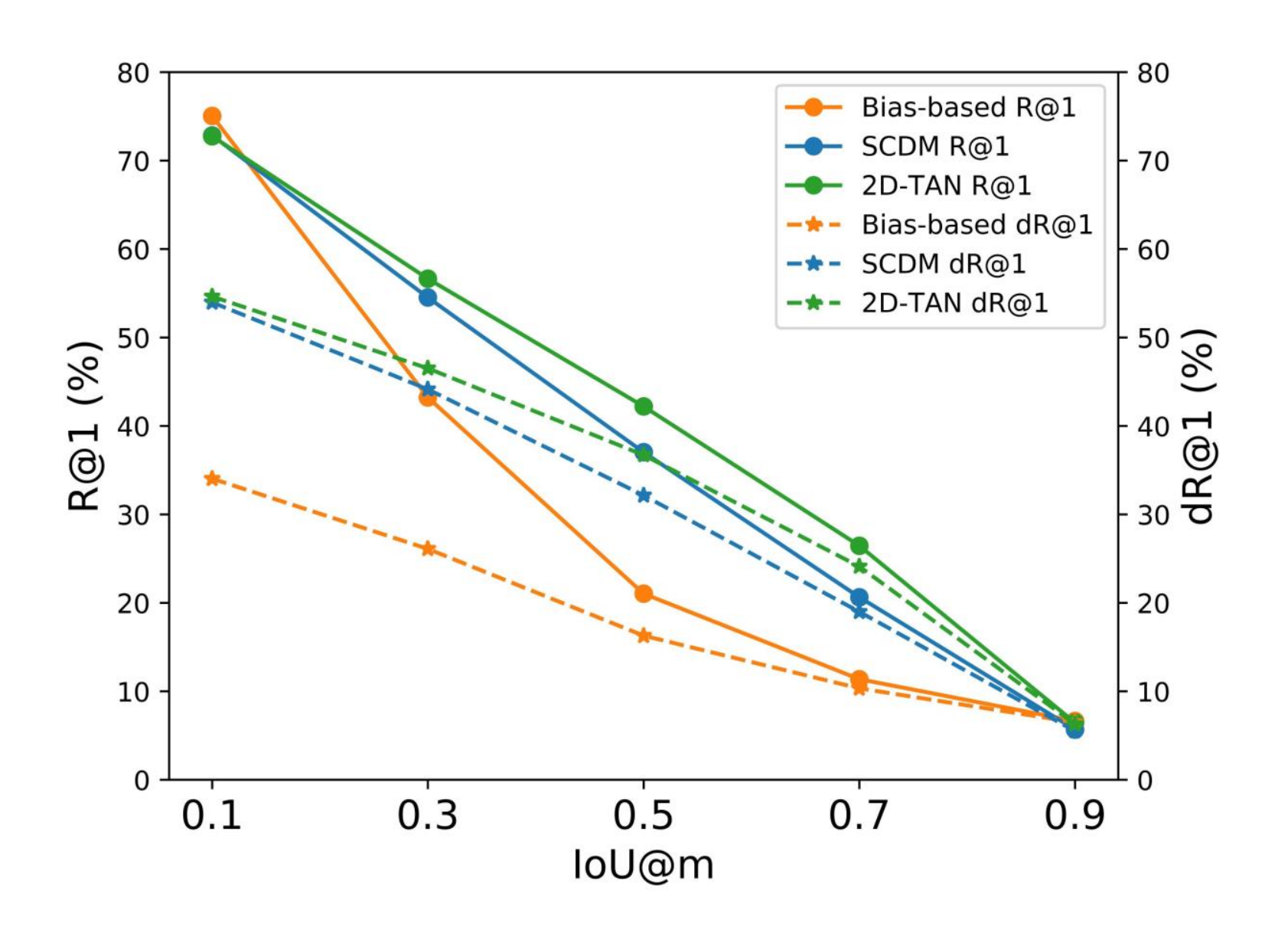}
	\caption{Performance (\%) comparisons of SOTA TSGV methods between original metric (R@$1$,IoU@$m$) and proposed metric (dR@$1$,IoU@$m$). All results come from the test set of ActivityNet Captions.}
	\label{fig:recall_compare}
\end{figure}

We report the performance of all mentioned TSGV methods with our proposed metric dR@$1$,IoU@$m$ in Table~\ref{tab:performance_drecall}. The trend of performance drop on the test-ood set compared to the test-iid set in Table~\ref{tab:performance_drecall} is similar to that in Figure~\ref{fig:ood_origin_compare}. These results verify again that current TSGV methods suffer from severe temporal annotation biases in the datasets, and fail to generalize to the OOD testing. Meanwhile, by comparing Table~\ref{tab:performance_drecall} and Figure~\ref{fig:ood_origin_compare}, we can observe that the dR@$1$,IoU@$m$ values are smaller than the R@$1$,IoU@$m$ values. For example, the SCDM model achieves score 32.5\% in R@$1$,IoU@$0.7$ while score 30.8\% in dR@$1$,IoU@$0.7$ on the test-iid set of Charades-CD. Such phenomenon is adhere to our definition of dR@$1$,IoU@$m$. For more clearer illustration, we further compare the dR@$1$ and R@$1$ scores under different IoUs of some SOTA methods in Figure~\ref{fig:recall_compare}. When the IoU threshold is small, dR@$1$ is much lower than R@$1$, and the gap between them gradually decreases with the increase of IoU threshold. Interestingly, we find that the naive Bias-based baseline achieves even better results than SCDM and 2D-TAN methods in the R@$1$,IoU@$0.1$ metric, while reversely in the dR@$1$,IoU@$0.1$ metric. These results indicate that recall values under small IoU thresholds are unreliable and overrated: although some moment predictions meet the IoU requirement, they still have a great discrepancy to the ground-truth moments. Instead, our proposed dR@$n$,IoU@$m$ metric can alleviate this problem since it can discount the recall value based on the temporal distance between the predicted and ground-truth moment temporal locations. When the prediction meets the larger IoU requirements, the discount will be smaller, \ie, the dR@$n$,IoU@$m$ values and R@$n$,IoU@$m$ values will be closer to each other. Therefore, our predicted dR@$n$,IoU@$m$ metric is more stable on different IoU thresholds, and it can suppress some inflating results (such as Bias-based or PredictAll baselines) caused by the moment annotation biases in the datasets. Meanwhile, these results further reveal that it is more reliable to report the grounding accuracy on large IoUs.

\section{Conclusion}

In this paper, we take a closer look at the existing evaluation protocol of the Temporal Sentence Grounding in Videos (TSGV) task, and we find that both the prevailing dataset splits and evaluation metric are the devils to cause the unreliable benchmarking: the datasets have obvious moment annotation biases and the metric is prone to overrating the model performance. To solve these problems, we propose to re-split the current Charades-STA and ActivityNet Captions datasets by making the ground-truth moment annotation distributions different in the training and test set. Meanwhile, we propose a new evaluation metric to alleviate the inflating evaluations caused by dataset annotation biases such as overlong ground-truth moments. The proposed data splits and metric serve as a promising test-bed to monitor the progress in TSGV. We also thoroughly evaluate eight state-of-the-art TSGV methods with the new evaluation protocol, opening the door for future research.

\section{Acknowledgments}
This work was supported by the National Key Research and Development Program of China under Grant No.2020AAA0106301, National Natural Science Foundation of China No.62050110 and Tsinghua GuoQiang Research Center Grant 2020GQG1014.

%
%
\balance
\bibliographystyle{ACM-Reference-Format}
\bibliography{sample-base}
\end{document}